\documentclass{bmvc2k}

\usepackage{multirow}
\usepackage{amsmath}
\usepackage{amssymb}
\usepackage{marvosym}

\let\svthefootnote\thefootnote
\newcommand\freefootnote[1]{%
  \let\thefootnote\relax%
  \footnotetext{#1}%
  \let\thefootnote\svthefootnote%
}

\title{MicroBT: Centroid-Guided Synergistic Learning for Multi-Modal Micro Brain Tumor Segmentation and Counting}

\addauthor{Zhenye Lou}{leonlou0921@gmail.com}{1\textsuperscript{*}}
\addauthor{Qing Xu}{scxqx1@nottingham.ac.uk}{2\textsuperscript{*}}
\addauthor{Stefanie Thust}{stefanie.thust@nottingham.ac.uk}{3}
\addauthor{Dorothee P. Auer}{dorothee.auer@nottingham.ac.uk}{3}
\addauthor{Zhen Chen}{z.chen@polyu.edu.hk}{4\textsuperscript{(\Letter)}}
\addauthor{Wenting Duan}{wduan@lincoln.ac.uk}{1}

\addinstitution{
 School of Engineering and Physical Science,\\
 University of Lincoln,\\
 Lincoln, UK
}
\addinstitution{
 School of Computer Science,\\
 University of Nottingham,\\
 Nottingham, UK
}
\addinstitution{
 Queen’s Medical Centre,\\
 University of Nottingham,\\
 Nottingham, UK
}
\addinstitution{
 Department of Data Science and Artificial Intelligence,\\
 The Hong Kong Polytechnic University,\\
 Hong Kong SAR
}

\runninghead{MicroBT}{Centroid-Guided Learning for Micro Brain Tumor}

\begin{document}

\maketitle
\begin{abstract}
Multi-modal brain tumor segmentation is essential for clinical diagnosis, radiotherapy planning, and longitudinal monitoring. While recent studies have achieved remarkable overall segmentation performance, they rely on overlap-based objectives that inherently favor large regions, causing sub-3 mm micro-lesions to be systematically suppressed during training. Moreover, existing pixel-level segmentation frameworks lack explicit instance-level modeling, leading to unreliable lesion counting that directly impacts clinical staging and therapeutic decisions. To address these challenges, we propose \textbf{MicroBT}, a unified multi-modal framework for micro brain tumor segmentation and counting. Our core insight is that a fundamental geometric property is inherently scale-invariant: \textit{regardless of size, each lesion possesses a well-defined centroid}. Leveraging this, we first devise a centroid-anchored synergistic paradigm that utilizes predicted centroids as shared anchors to simultaneously condition dense segmentation and instance-level localization. To counteract the tumor size imbalance, we further devise a Volume-Calibrated Loss (VC-Loss) that adaptively amplifies the gradient contribution of micro-lesions, alongside a Size-Aware Curriculum Optimization (SACO) that stabilizes multi-task training by establishing global representations before progressively unlocking fine-grained localization. Extensive experiments demonstrate that MicroBT outperforms state-of-the-art methods, particularly on metastases, where micro-lesion ($\leq$3\,mm) localization precision improves by $\sim$7\% and counting estimation errors decrease by $\sim$26\%. The code is available at https://github.com/leonlou921/MicroBT. \freefootnote{$*$ Equal Contribution, \Letter \, Corresponding Author.} 
\end{abstract}

\section{Introduction}

Multi-modal brain tumor segmentation is a representative task in 3D medical dense prediction, requiring joint exploitation of complementary MRI sequences (\textit{e.g.}, T1, T1c, T2, and FLAIR) for accurate volumetric delineation \cite{menze2014multimodal,hosseini2024topk,xu2023dcsau}. This task holds clinical significance for diagnosis and radiotherapy planning \cite{topff2025data,ali2025attention,dasgupta2025omnisegnet,kolahi2024msa2net}, while posing fundamental vision challenges, including multi-modal fusion, extreme scale variation, and unified dense-to-instance prediction, making it a highly active challenge in 3D medical image analysis.

The classical methods for multi-modal brain tumor segmentation built upon 3D convolutional architectures, \textit{e.g.}, 3D U-Net variants \cite{isensee2021nnu,milletari2016v,yao2025brats,dang2024singr}, leveraging hierarchical feature extraction to delineate tumor sub-regions, yet such convolution-based designs have limited capacity to model long-range spatial dependencies across heterogeneous tumor morphology. To address this, Transformer-based methods \cite{chen2024transunet,shaker2024unetr++,zhou2023nnformer,luo2025msm} introduced self-attention mechanisms that capture global context across volumetric inputs, enabling more holistic scene understanding. Moreover, recent state-space-based architectures \cite{liu2024swin,xing2025segmamba,ma2024u,gong2025nnmamba} have demonstrated the potential of efficient long-range dependency modeling.

\begin{figure*}[!t]
\centering
\includegraphics[width=0.8\textwidth]{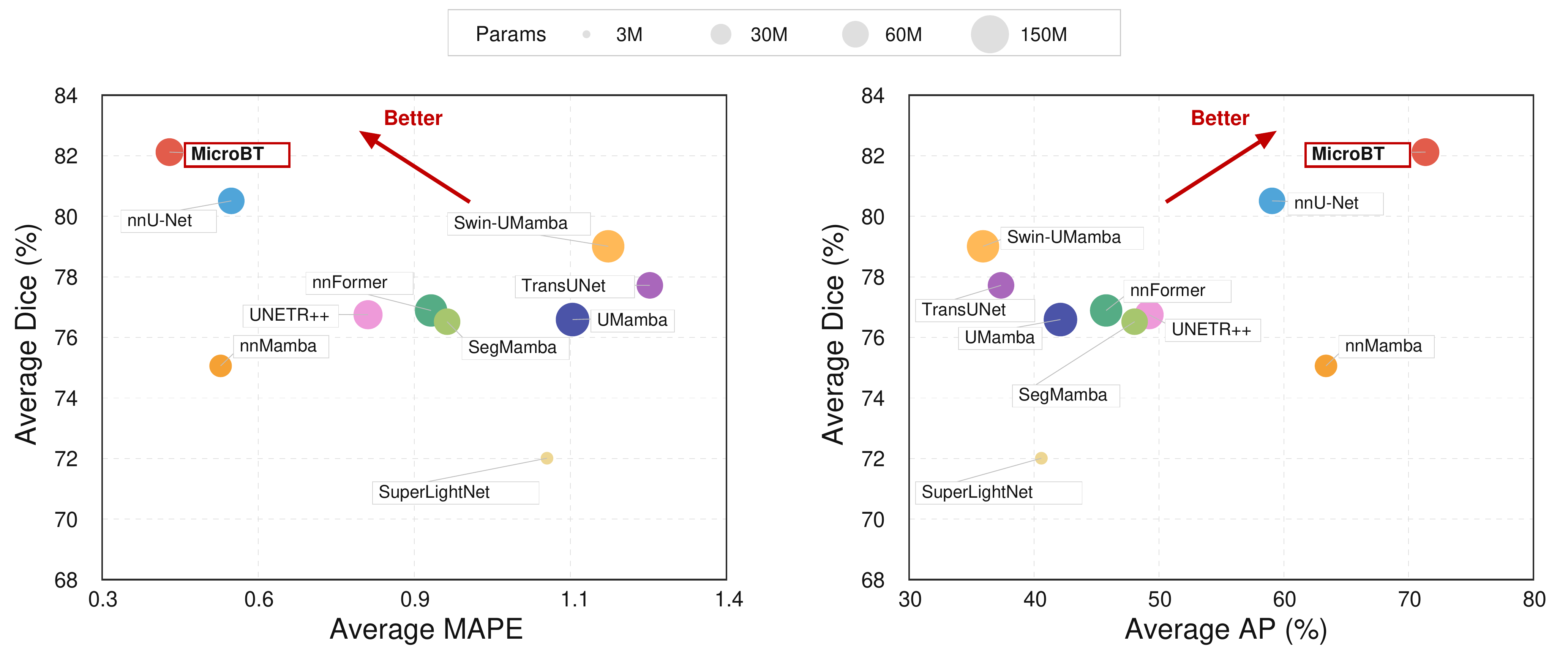}
\caption{Comparison of segmentation quality and counting reliability across four BraTS datasets. Our MicroBT framework achieves the best segmentation accuracy and counting performance with comparable computational cost to existing methods, demonstrating that centroid-guided synergistic learning effectively bridges dense segmentation and instance-level localization for multi-modal micro brain tumors.}
\label{fig:scatter}
\end{figure*}

Despite these advancements, existing methods \cite{liu2024swin,xing2025segmamba,chen2024transunet} rely on overlap-based objectives that implicitly favor large regions, systematically suppressing sub-3 mm micro-lesions, and lack explicit instance-level modeling, leading to unreliable lesion counting. In fact, micro-lesions smaller than 3 mm are of paramount clinical significance in brain metastasis and multicentric tumor management, as they may signal early dissemination, treatment-resistant foci, or emerging recurrence, and their accurate enumeration directly influences staging and therapeutic decisions \cite{topff2025data}. For instance, in stereotactic radiosurgery planning, missing even a single micro-lesion can lead to inadequate dose coverage and subsequent disease progression \cite{tanzhu2024metastatic}. This critical clinical demand motivates us to develop a unified framework for micro brain tumor segmentation and counting.

To address this, we propose \textbf{MicroBT}, a unified multi-modal framework for micro brain tumor segmentation and counting. Our core insight is that a fundamental geometric property is inherently scale-invariant: \textit{regardless of size, each lesion possesses a well-defined centroid}. Leveraging this, we first devise a Centroid-Anchored Synergistic (CAS) paradigm that utilizes predicted centroids as shared anchors to simultaneously condition dense segmentation and instance-level localization within a single architecture. To counteract the tumor size imbalance, we further introduce a Volume-Calibrated Loss (VC-Loss) that adaptively amplifies the gradient contribution of micro-lesions, alongside a Size-Aware Curriculum Optimization (SACO) that stabilizes multi-task training by establishing global representations before progressively unlocking fine-grained localization. To the best of our knowledge, MicroBT is the first framework dedicated to micro brain tumor segmentation and counting in multi-modal MRI. As illustrated in Fig.~\ref{fig:scatter}, extensive experiments across four public BraTS challenge datasets covering gliomas, meningiomas, and metastases consistently demonstrate that MicroBT achieves superior performance across all three tumor types, particularly improving average Dice by 1.61\% over nnU-Net~\cite{isensee2021nnu} while reducing Mean Absolute Percentage Error (MAPE) by 18.3\% and boosting localization precision (AP$_{50}$) by 12.29\%. The contributions of this work are summarized as follows:
\begin{itemize}
\item We propose MicroBT, a centroid-guided synergistic learning framework for micro brain tumor segmentation and counting that leverages lesion centroids as scale-invariant anchors to overcome the suppression of micro-lesions by overlap-based objectives.
\item We design a CAS that generates lesion centroids as scale-invariant anchors, and a centroid-guided dual-branch decoder that uses these anchors to jointly condition dense segmentation and instance-level localization within a unified architecture.
\item We devise a VC-Loss that adaptively amplifies the gradient contribution of micro-lesions, together with a SACO strategy that stabilizes multi-task training by establishing global representations before progressively unlocking fine-grained localization.
\item We conduct extensive experiments on four public BraTS challenge datasets, and our MicroBT outperforms state-of-the-art methods on segmentation and counting tasks across gliomas, meningiomas, and metastases.
\end{itemize}

\section{Related Work}

\subsection{Multi-Modal Brain Tumor Segmentation}
\label{sec:rw_brats}

Brain tumors, encompassing gliomas, meningiomas, and metastases, are among the most prevalent malignancies of the central nervous system, where accurate delineation from multi-modal MRI~\cite{luo2025msm} directly informs surgical planning, radiotherapy targeting, and longitudinal treatment assessment. Early methods~\cite{isensee2021nnu, milletari2016v,liu2025harnessing,reich2021oss} adopted 3D U-shaped CNNs to delineate tumor sub-regions, but their limited receptive field constrained long-range spatial dependency modeling across heterogeneous tumor morphology. Subsequent Transformer-based methods~\cite{wang2021transbts,chen2024transunet,zhou2023nnformer,shaker2024unetr++,lin2023ckd,she2023eoformer} introduced self-attention to capture global volumetric context, while recent state-space-based architectures~\cite{liu2024swin,xing2025segmamba,ma2024u,gong2025nnmamba,jiao2026tm} achieved efficient long-range modeling with linear complexity. Lightweight and efficiency-oriented designs~\cite{yu2025superlightnet,chen2026switchnet} have further pushed practical deployment boundaries. Despite these advances, existing methods uniformly treat tumors of all scales as a single optimization target under overlap-based objectives, systematically suppressing sub-3\,mm micro-lesions and lacking explicit instance-level modeling for reliable lesion counting. In contrast, our MicroBT introduces a centroid-anchored synergistic paradigm that preserves micro-lesion sensitivity and unifies dense segmentation with instance-level localization within a single framework.

\subsection{Brain Metastases Analysis}
\label{sec:rw_mets}

Brain metastases are the most common intracranial malignancies in adults, occurring approximately ten times more frequently than primary brain tumors~\cite{van2025response}, with lesion count serving as a key criterion for treatment selection between whole-brain radiotherapy and stereotactic radiosurgery~\cite{tanzhu2024metastatic} and longitudinal monitoring~\cite{aizer2026treatment}. Early deep learning approaches~\cite{xue2020deep} adopted cascaded 3D fully convolutional networks for automatic detection and segmentation, while subsequent works improved sensitivity to small metastases through multi-scale classification~\cite{li2023mri} and high-resolution feature preservation~\cite{qu2023construction}. Recent studies further validated clinical efficacy in radiologist-assisted contouring~\cite{luo2024automated} and longitudinal monitoring of stereotactic radiotherapy outcomes~\cite{jalalifar2023automatic}. However, existing approaches typically treat detection, segmentation, and counting as separate tasks, lacking a unified framework that simultaneously optimizes segmentation accuracy and counting reliability for micro-lesions. In contrast, our MicroBT unifies dense segmentation and instance-level localization through centroid-guided synergistic learning, with the volume-calibrated loss and size-aware curriculum specifically designed to preserve micro-lesion sensitivity.

\begin{figure}[!t]
\centering
\includegraphics[width=0.8\textwidth]{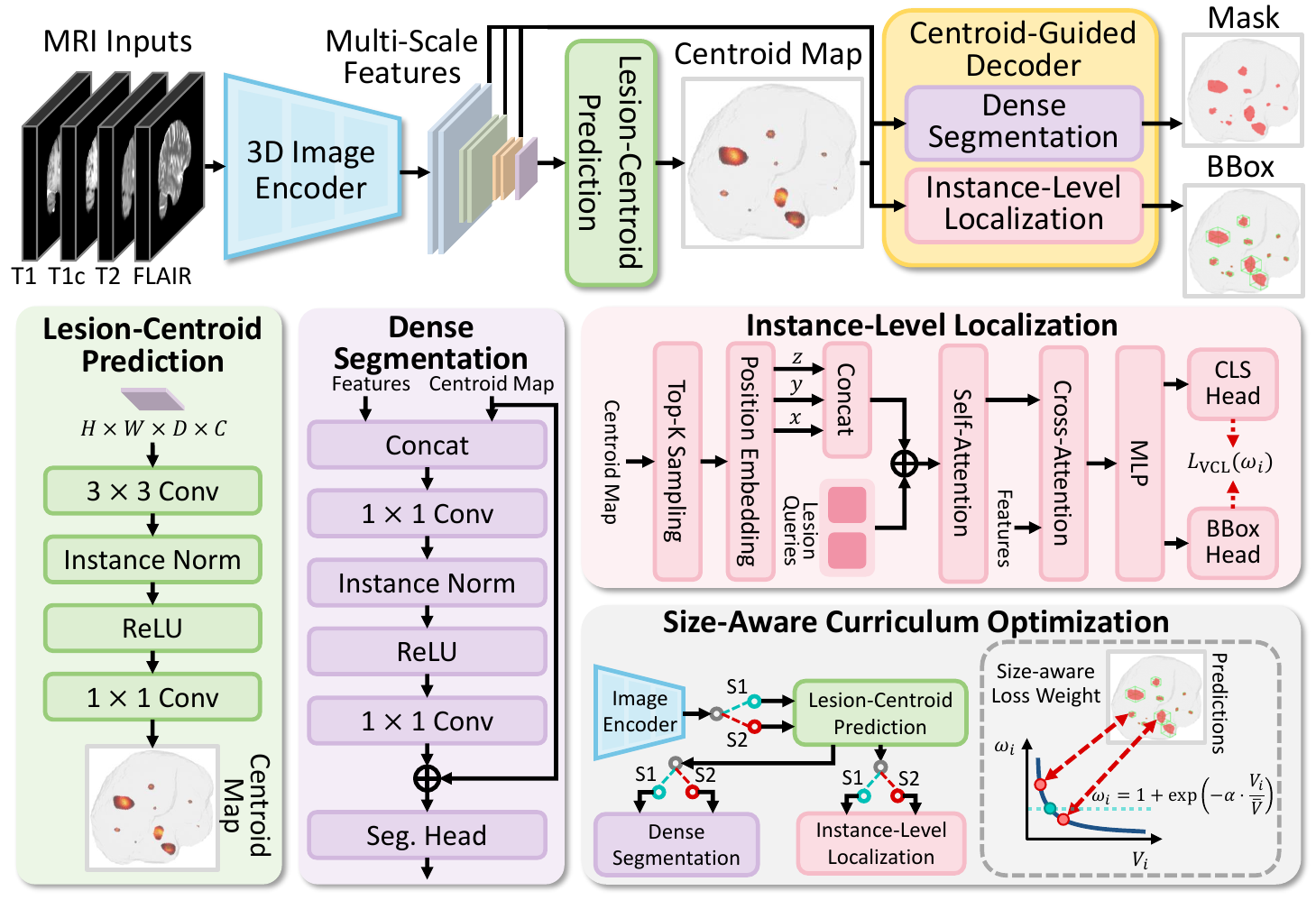}
\caption{The overview of our MicroBT framework for multi-modal micro brain tumor segmentation and counting. MicroBT utilizes predicted centroids as shared anchors to simultaneously condition dense segmentation and instance-level localization.}
\label{fig:overview}
\end{figure}

\section{Methodology}

As illustrated in Fig.~\ref{fig:overview}, we devise the MicroBT framework with a centroid-guided synergistic learning paradigm for multi-modal micro brain tumor segmentation and counting. To accomplish this, we utilize predicted centroids as shared anchors to simultaneously condition dense segmentation and instance-level localization, devise a VC-Loss to adaptively amplify the gradient contribution of micro-lesions, and a SACO to stabilize multi-task training by establishing global representations before progressively unlocking fine-grained localization.

\subsection{Centroid-Anchored Synergistic Paradigm}
\label{sec:paradigm}

Existing multi-modal brain tumor segmentation methods~\cite{milletari2016v,isensee2021nnu} rely on overlap-based objectives that implicitly favor large regions, systematically suppressing sub-3\,mm micro-lesions, and lack explicit instance-level modeling, leading to unreliable lesion counting. To address this issue, we propose the Centroid-Anchored Synergistic paradigm (CAS) that introduces a scale-invariant latent variable as a shared anchor to enable synergistic optimization between dense segmentation and instance-level localization. Specifically, our goal is to train a model $f_\Theta: x \rightarrow \{y_{\rm seg}, y_{\rm det}\}$, where $\Theta = \{\theta_{\rm ctr}, \theta_{\rm seg}, \theta_{\rm det}\}$ represents the parameters for centroid prediction $\theta_{\rm ctr}$, dense segmentation $\theta_{\rm seg}$, and instance-level localization $\theta_{\rm det}$, with $x \in \mathbb{R}^{C \times D \times H \times W}$ being the volumetric multi-modal MRI input, $y_{\rm seg} \in \mathbb{R}^{D \times H \times W}$ the pixel-wise segmentation map, and $y_{\rm det} = \{(c_i, b_i)\}_{i=1}^M$ the instance-level detection set. Our core insight is that a fundamental geometric property is inherently scale-invariant: \textit{regardless of size, each lesion possesses a well-defined centroid}~\cite{duan2019centernet}. Leveraging this property, we introduce a centroid heatmap $z_{\rm ctr}$ as the shared latent anchor that acts as a sufficient statistic decoupling the two downstream tasks, \textit{i.e.}, $p(y_{\rm seg}, y_{\rm det} \mid x, z_{\rm ctr}) = p(y_{\rm seg} \mid x, z_{\rm ctr}) \cdot p(y_{\rm det} \mid x, z_{\rm ctr})$. Under this conditional independence, the joint distribution factorizes as:
\begin{equation}
\label{eq:paradigm_joint}
  p(z_{\rm ctr}, y_{\rm seg}, y_{\rm det} \mid x; \Theta) = p(z_{\rm ctr} \mid x; \theta_{\rm ctr}) \cdot p(y_{\rm seg} \mid x, z_{\rm ctr}; \theta_{\rm seg}) \cdot p(y_{\rm det} \mid x, z_{\rm ctr}; \theta_{\rm det}).
\end{equation}
Eq.~\eqref{eq:paradigm_joint} reveals the centroid-anchored synergy through a hierarchical conditional factorization, where the shared latent anchor $z_{\rm ctr}$ serves three complementary roles: $p(z_{\rm ctr} \mid x; \theta_{\rm ctr})$ captures the scale-invariant geometric prior, providing a robust anchor equally informative for micro-lesions and large tumors; $p(y_{\rm seg} \mid x, z_{\rm ctr}; \theta_{\rm seg})$ conditions the segmentation decoder on this anchor, injecting an explicit spatial prior that preserves micro-lesion representations; and $p(y_{\rm det} \mid x, z_{\rm ctr}; \theta_{\rm det})$ conditions the detection decoder on the same anchor, replacing blind volumetric search with targeted query initialization. Notably, $\theta_{\rm ctr}$ receives gradients not only from its own supervision, but also from both downstream tasks back-propagated through $z_{\rm ctr}$. This forces the centroid head to refine the anchor in a way that maximally benefits both dense segmentation and instance-level localization, while the downstream branches in turn enjoy a continuously sharpened spatial prior. In this way, the proposed CAS establishes a shared latent anchor that simultaneously conditions dense segmentation and instance-level localization, counteracting the systematic suppression of micro-lesions.

\subsection{The MicroBT Architecture Designs}
\label{sec:architecture}

Following the centroid-anchored synergistic paradigm, we design the MicroBT architecture to operationalize the conditional chain in Eq.~\eqref{eq:paradigm_joint}. As illustrated in Fig.~\ref{fig:overview}, MicroBT consists of a 3D CNN image encoder ~\cite{isensee2021nnu}, a lesion-centroid prediction module, and a centroid-guided decoder with two parallel branches for dense segmentation and instance-level localization. The encoder processes the input $X$ to extract multi-scale features $F = \{f_1, f_2, \ldots, f_L\}$, which are then distributed to the downstream modules.

\noindent\textbf{Lesion-Centroid Prediction.} From the penultimate encoder feature $f_{L-1}$, which retains sufficient spatial resolution for localizing micro-lesions, a lightweight centroid head predicts the centroid heatmap:
\begin{equation}
\label{eq:heatmap}
  H = \sigma\!\left(\mathrm{Conv}_{1\times1}\!\left(
    \mathrm{ReLU}\!\left(
      \mathrm{IN}\!\left(
        \mathrm{Conv}_{3\times3}(f_{L-1})
      \right)
    \right)
  \right)\right),
\end{equation}
where $\mathrm{IN}$ denotes Instance Normalization and $\sigma$ is the sigmoid function. The predicted centroid map $H$ then serves as the shared anchor to simultaneously condition the two downstream branches within the centroid-guided decoder.

\noindent\textbf{Dense Segmentation.} To counteract the tendency of overlap-based objectives to overlook micro-lesions, we inject $H$ as a spatial attention prior into the skip-connection pathway via a residual feature injector. Specifically, $f_{L-1}$ is concatenated with $H$ and projected back to the original channel dimension:
\begin{equation}
  f'_{L-1} = f_{L-1} \oplus \mathrm{Conv}_{1\times1}\!\left(
    \mathrm{ReLU}\!\left(
      \mathrm{IN}\!\left(
        \mathrm{Conv}_{1\times1}([\,f_{L-1};\; H\,])
      \right)
    \right)
  \right),
\end{equation}
where $[\,\cdot\,;\,\cdot\,]$ denotes channel-wise concatenation. The enhanced $f'_{L-1}$ replaces $f_{L-1}$ in the skip connections fed to the segmentation head, yielding $\hat{Y}_{\rm seg} = D_{\rm seg}(\{f_1,\ldots,f_{L-2},f'_{L-1},f_L\})$.

\noindent\textbf{Instance-Level Localization.} To replace blind volumetric search with targeted localization, we perform Top-$K$ sampling on the centroid map to obtain $K$ candidate centroid coordinates $\mathcal{P} = \{(z_k, y_k, x_k)\}_{k=1}^K$, where each triplet denotes the spatial position of a detected peak in $H$, and encode them via 3D sinusoidal positional embeddings~\cite{vaswani2017attention} to form lesion queries:
\begin{equation}
  Q_{\rm pos} = \mathrm{Linear}\!\left(
    \bigl[\,
      \mathrm{SinEmb}(z_k);\;
      \mathrm{SinEmb}(y_k);\;
      \mathrm{SinEmb}(x_k)
    \,\bigr]_{k=1}^K
  \right),
\end{equation}
where $\mathrm{SinEmb}(\cdot)$ denotes the sinusoidal positional encoding applied independently to each spatial axis and $\mathrm{Linear}(\cdot)$ is a learnable projection that maps the concatenated axis embeddings to the query dimension. The lesion queries are combined with learnable content embeddings and refined through self-attention and cross-attention layers against the projected feature memory. A CLS head and a BBox head then predict the class probability and normalized bounding box for each query, respectively. In this way, the centroid anchor bridges dense segmentation and instance-level localization within a unified architecture.

\subsection{Volume-Calibrated Loss}
\label{sec:vcloss}

The classical methods \cite{milletari2016v,isensee2021nnu} rely on overlap-based objectives that implicitly favor large regions, systematically suppressing sub-3\,mm micro-lesions. This bias arises because both overlap-based segmentation losses and bbox regression losses scale with absolute lesion dimensions: micro-lesions occupying only a few voxels contribute disproportionately small gradients compared with large tumors, regardless of their relative prediction error. To counteract the tumor size imbalance, we devise the VC-Loss that adaptively amplifies the gradient contribution of micro-lesions. Specifically, for each ground-truth instance with volume $V_i$, we define a calibration weight:
\begin{equation}
\label{eq:vcloss}
  \omega_i = 1 + \exp\!\left(-\alpha \cdot \frac{V_i}{\bar{V}}\right),
\end{equation}
where $\bar{V}$ is the mean volume across all instances in the current batch and $\alpha$ is a scaling factor. This formulation ensures that micro-lesions with $V_i \ll \bar{V}$ receive amplified gradients, while large tumors converge to baseline weighting, achieving smooth calibration without abrupt thresholds. Three properties make this formulation principled. First, the exponential form bounds $\omega_i$ within $(1, 2]$, ensuring stable amplification without gradient explosion. Second, the batch-adaptive normalization by $\bar{V}$ removes the need for any global size threshold, allowing the calibration to adapt automatically to the lesion size distribution of each sample. Third, the weight is applied selectively to per-instance bbox regression, preserving the standard balance among the segmentation, centroid, and classification objectives. The VC-Loss is defined by:
\begin{equation}
\label{eq:vcl}
  L_{\text{VC-Loss}} = L_{\rm seg} + L_{\rm ctr} + \frac{1}{M} \sum_{i=1}^{M} \left[
    L_{\rm cls}^{(i)} + \omega_i \left(
    \| v_i - \hat{v}_{\sigma^*(i)} \|_1
    + L_{\rm GIoU}(v_i, \hat{v}_{\sigma^*(i)})
    \right)
  \right],
\end{equation}
where $L_{\rm seg}$ combines the Dice loss and Cross Entropy loss for segmentation, $L_{\rm ctr}$ is the MSE loss between the predicted centroid heatmap $H$ and the Gaussian-smoothed ground-truth map, $v_i$ and $\hat{v}_{\sigma^*(i)}$ denote the ground-truth and predicted 3D volumetric bounding boxes under Hungarian matching~\cite{kuhn1955hungarian,carion2020end}, and $L_{\rm cls}$ is the focal loss~\cite{lin2017focal}. In this way, the proposed VC-Loss explicitly counteracts the systematic suppression of micro-lesions during optimization, ensuring that small but clinically meaningful lesions receive sufficient gradient signals for reliable segmentation and counting in micro brain tumor scenarios.

\subsection{Size-Aware Curriculum Optimization}
\label{sec:saco}

To stabilize multi-task training for MicroBT, we devise the SACO that establishes global representations before progressively unlocking fine-grained localization, following the curriculum learning paradigm~\cite{bengio2009curriculum}. The need for such a curriculum is amplified by the synergistic structure of CAS: since the centroid head receives back-propagated gradients from both downstream tasks through the shared anchor $z_{\rm ctr}$, any noisy signal from one branch immediately propagates to all other components. Directly activating the VC-Loss from the start exacerbates this risk: amplified micro-lesion gradients enter the optimization before the shared encoder has built reliable representations, producing unstable centroid predictions that in turn propagate errors back to both segmentation and detection branches. As illustrated in the SACO panel of Fig.~\ref{fig:overview}, we therefore introduce a two-stage curriculum aligned with the natural large-to-small learning order, where each learnable component is switched between an unweighted Stage~I and a size-calibrated Stage~II.

Concretely, in Stage~I, all parameters across the encoder, centroid head, and task-specific decoders are trained end-to-end with a uniform learning rate and size weights fixed at $\omega_i = 1$, allowing the shared encoder to build robust, scale-agnostic anatomical representations and a stable centroid anchor without interference from amplified micro-lesion gradients. In Stage~II, we activate the full VC-Loss with size-aware weighting $\omega_i$ and impose a differential learning rate schedule~\cite{loshchilov2017decoupled}, where the encoder is updated at a significantly attenuated rate while the task-specific heads continue at the original rate. This design allows the amplified micro-lesion gradients to drive the centroid and detection heads to sharpen fine-grained localization sensitivity, while the attenuated encoder preserves the global representations established in Stage~I, preventing catastrophic forgetting. In this way, SACO progressively transitions the optimization focus from global context modeling to micro-lesion-specific refinement, stabilizing the multi-task training of our MicroBT framework.

\section{Experiments}

\subsection{Datasets}
\label{sec:datasets}

To validate the effectiveness of the proposed MicroBT, we conduct experiments on four public BraTS challenge datasets covering three brain tumor types. For metastasis segmentation, we adopt BraTS-Lighthouse-MET~2025~\cite{maleki2025analysis}, a multi-institutional dataset of brain metastasis cases. For meningioma segmentation, we employ ASNR-MICCAI BraTS-MEN~2023~\cite{labella2023asnr}, a large-scale multi-institutional benchmark of intracranial meningiomas. For glioma segmentation, we combine BraTS-PED~2024~\cite{kazerooni2024brain} and BraTS-GLI~2025~\cite{de20242024}, two multi-institutional datasets covering pediatric and adult pre-treatment gliomas, respectively. All four datasets follow the BraTS protocol with four MRI modalities (T1, T1ce, T2, FLAIR), co-registered, skull-stripped, and resampled to $1\,\mathrm{mm}^3$ isotropic spacing. We retain cases containing at least one micro-lesion ($ \leq 3\,\mathrm{mm}$ maximum 3D diameter), resulting in 1341 cases verified by two certificated radiologists.

\begin{table*}[t]
\centering
\caption{Comparison with state-of-the-arts on Micro Brain Tumor
Segmentation.}
\tabcolsep=0.2cm
\label{tab:merged_results_v2}
\resizebox{\textwidth}{!}{%
\begin{tabular}{l|cc|cc|cc|cc}
\hline
\multirow{2}{*}{Method} & \multicolumn{2}{c|}{Metastases} & \multicolumn{2}{c|}{Meningiomas} & \multicolumn{2}{c|}{Gliomas} & \multicolumn{2}{c}{Average} \\
\cline{2-3} \cline{4-5} \cline{6-7} \cline{8-9}
 & Dice $\uparrow$ &HD95 $\downarrow$ & Dice $\uparrow$ &HD95 $\downarrow$ & Dice $\uparrow$ &HD95 $\downarrow$ & Dice $\uparrow$ &HD95 $\downarrow$ \\
\hline
nnU-Net~\cite{isensee2021nnu}          & 72.15 & 24.65 & 80.85 & 19.62 & 88.52 & 6.75 & 80.51 & 17.01 \\
TransUNet~\cite{chen2024transunet}  & 65.76 & 28.16 & 79.11 & 26.62 & 88.28 & 8.76 & 77.72 & 21.18 \\
nnFormer~\cite{zhou2023nnformer}       & 67.39 & 26.10 & 76.36 & 21.82 & 86.91 & 9.02 & 76.89 & 18.98 \\
Swin-UMamba~\cite{liu2024swin}  & 68.85 & 19.45 & 79.82 & 15.85 & 88.35 & 6.65 & 79.01 & 13.98 \\
UMamba~\cite{ma2024u}           & 64.07 & 29.93 & 78.14 & 30.83 & 87.55 & 10.55 & 76.59 & 23.77 \\
UNETR++~\cite{shaker2024unetr++}         & 64.79 & 24.33 & 78.39 & 16.73 & 87.07 & 11.42 & 76.75 & 17.49 \\
SegMamba~\cite{xing2025segmamba}       & 63.45 & 28.45 & 77.85 & 15.75 & 88.25 & 6.95 & 76.52 & 17.05 \\
nnMamba~\cite{gong2025nnmamba}         & 64.83 & 20.31 & 74.78 & 17.92 & 85.56 & 7.15 & 75.06 & 15.13 \\ 
SuperLightNet~\cite{yu2025superlightnet} & 60.15 & 32.50 & 70.13 & 24.44 & 85.74 & 8.08 & 72.01 & 21.67 \\
\hline
MicroBT (Ours)        & \textbf{74.68} & \textbf{18.98} & \textbf{82.45} & \textbf{15.47} & \textbf{89.24} & \textbf{6.48} & \textbf{82.12} & \textbf{13.64} \\
\hline
\end{tabular}%
}
\end{table*}

\begin{table*}[!t]
\centering
\caption{Comparison with state-of-the-arts on Micro Brain Tumor
Counting.}
\tabcolsep=0.13cm
\label{tab:detection_counting_results}
\resizebox{\textwidth}{!}{%
\begin{tabular}{l|cc|cc|cc|cc}
\hline
\multirow{2}{*}{Method} & \multicolumn{2}{c|}{Metastases} & \multicolumn{2}{c|}{Meningiomas} & \multicolumn{2}{c|}{Gliomas} & \multicolumn{2}{c}{Average} \\
\cline{2-3} \cline{4-5} \cline{6-7} \cline{8-9}
 & MAPE $\downarrow$ & AP$_{\rm 50}$ $\uparrow$ & MAPE $\downarrow$ & AP$_{\rm 50}$ $\uparrow$ & MAPE $\downarrow$ & AP$_{\rm 50}$ $\uparrow$ & MAPE $\downarrow$ & AP$_{\rm 50}$ $\uparrow$ \\
\hline
nnDetection~\cite{baumgartner2021nndetection} & 0.852 & 25.58 & 1.865 & 38.55 & 1.017 & 39.53 & 1.245 & 34.55 \\
nnU-Net~\cite{isensee2021nnu}          & 0.568 & 47.51 & 0.536 & 64.18 & 0.598 & 65.49 & 0.567 & 59.06 \\
TransUNet~\cite{chen2024transunet}  & 0.907 & 30.46 & 1.894 & 33.94 & 1.011 & 47.66 & 1.271 & 37.35 \\
nnFormer~\cite{zhou2023nnformer}       & 0.701 & 48.66 & 0.992 & 42.64 & 1.015 & 46.01 & 0.903 & 45.77 \\
Swin-UMamba~\cite{liu2024swin}  & 0.773 & 28.36 & 1.812 & 41.33 & 1.019 & 38.04 & 1.201 & 35.91 \\
UMamba~\cite{ma2024u}           & 0.730 & 40.59 & 1.760 & 45.55 & 0.933 & 40.19 & 1.141 & 42.11 \\
UNETR++~\cite{shaker2024unetr++}         & 0.632 & 44.55 & 0.871 & 54.31 & 0.887 & 48.78 & 0.797 & 49.21 \\
SegMamba~\cite{xing2025segmamba}       & 0.707 & 38.89 & 1.269 & 54.70 & 0.814 & 50.57 & 0.930 & 48.05 \\
nnMamba~\cite{gong2025nnmamba}         & 0.602 & 58.65 & 0.408 & 62.36 & 0.638 & 69.13 & 0.549 & 63.38 \\
SuperLightNet~\cite{yu2025superlightnet} & 0.733 & 22.65 & 1.840 & 49.97 & 0.721 & 49.11 & 1.098 & 40.58 \\
\hline
MicroBT (Ours)        & \textbf{0.423} & \textbf{65.96} & \textbf{0.401} & \textbf{68.52} & \textbf{0.565} & \textbf{79.56} & \textbf{0.463} & \textbf{71.35} \\
\hline
\end{tabular}%
}
\end{table*}

\begin{figure}[!t]
\centering
\includegraphics[width=0.9\textwidth]{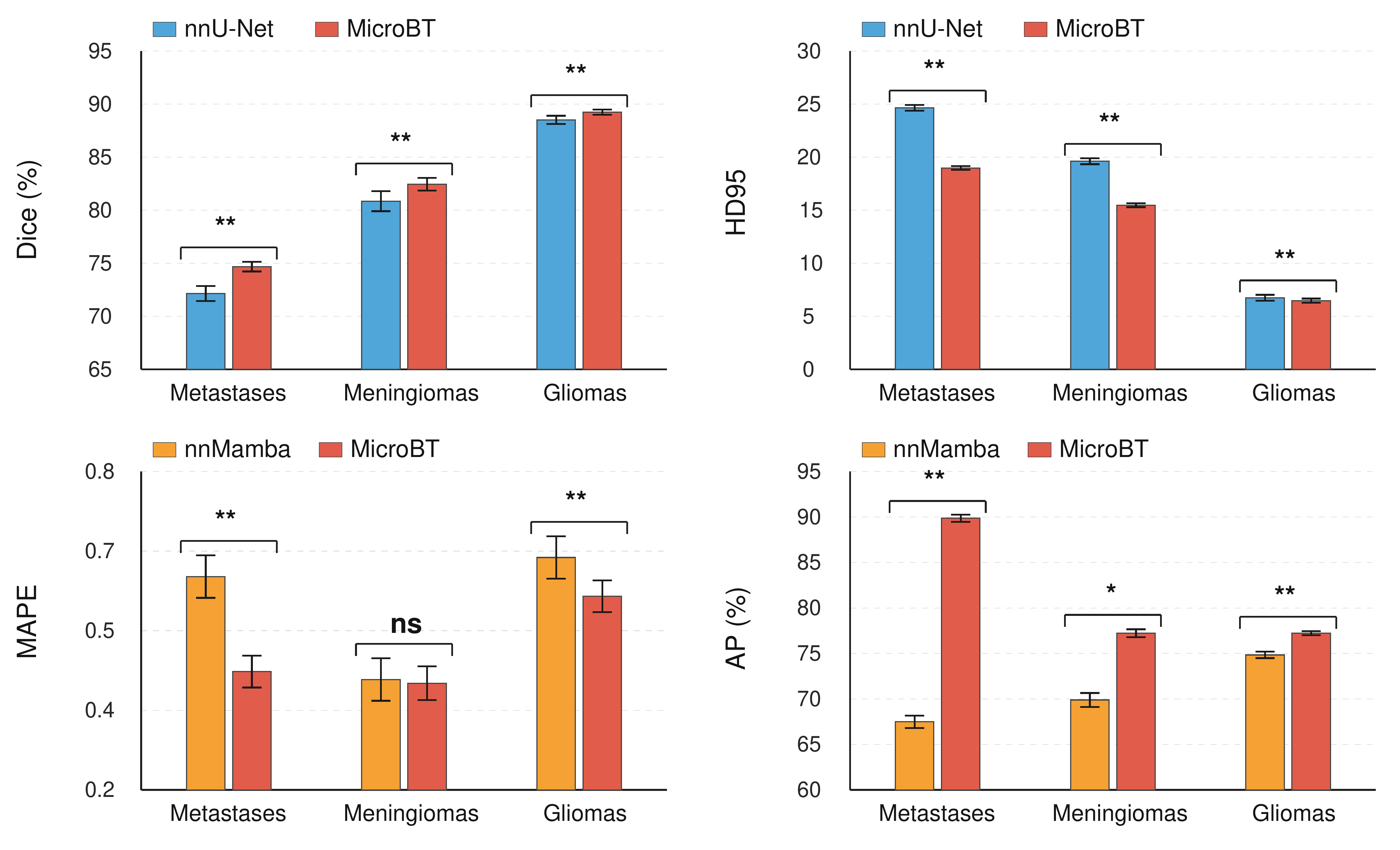}
\caption{Statistical significance analysis via paired $t$-tests across all test cases. Top: MicroBT vs.\ nnU-Net~\cite{isensee2021nnu} on segmentation metrics (Dice, HD95). Bottom: MicroBT vs.\ nnMamba~\cite{gong2025nnmamba} on counting metrics (MAPE, AP$_{50}$). **: $p < 0.01$, *: $p < 0.05$, ns: not significant. These results demonstrate that our MicroBT framework achieves significant improvements on nearly all metrics across three tumor types.}
\label{fig:significance}
\end{figure}

\subsection{Implementation Details}
\label{sec:implementation}

We implement MicroBT in PyTorch and conduct all experiments on a single NVIDIA RTX 3060 GPU. Each input volume is intensity-normalized per modality via z-score normalization, and randomly cropped to a fixed patch size of $128 \times 128 \times 128$ during training. Standard 3D augmentations, including random flipping, rotation, and intensity scaling, are applied to mitigate overfitting. We adopt a random 8:2 case-level split for training and testing within each dataset, with strict case-level separation enforced to prevent any test-set leakage. Data preprocessing follows the standard nnUNet~\cite{isensee2021nnu} protocol. The number of object queries is set to $K=50$, the Top-K is set to $q=50$, and the scaling factor in the VC-Loss (Eq.~\ref{eq:vcloss}) is set to $\alpha=0.5$. The network is optimized with AdamW~\cite{loshchilov2017decoupled} using a weight decay of $1\times10^{-2}$ and an initial learning rate of $1\times10^{-4}$. Following the proposed SACO strategy, Stage~I trains all parameters end-to-end for 300 epochs with a uniform learning rate and uniform size weighting ($\omega_i=1$). Stage~II then activates the full VC-Loss with size-aware weighting and attenuates the encoder learning rate to $1\times10^{-6}$ while keeping the task-specific heads at $1\times10^{-4}$, for an additional 200 epochs until convergence. At inference, we follow \cite{isensee2021nnu}, applying sliding-window prediction with an overlap ratio of 0.5, aggregating the per-window centroid heatmaps and bounding-box predictions, and applying non-maximum suppression at an IoU threshold of 0.2 to obtain the final per-case lesion count and instance localizations.

\subsection{Evaluation Metrics}
\label{sec:metrics}

To perform a comprehensive evaluation of segmentation quality, instance-level counting, and clinical detection reliability, we employ complementary metrics across three granularities. For segmentation, we report the Dice score for region-overlap fidelity and the 95\% Hausdorff Distance (HD95) for boundary accuracy. For instance-level counting and localization, we adopt the Mean Absolute Percentage Error (MAPE)~\cite{de2016mean}, which measures the average relative deviation between the predicted and ground-truth lesion counts per case, together with the Average Precision at an IoU threshold of $0.5$ (AP$_{50}$), which jointly captures lesion-level detection precision and recall under a strict 3D localization criterion. For clinically relevant detection analysis, following the evaluation protocol of~\cite{topff2025data}, we report object-level Free-Response Receiver Operating Characteristic (FROC) curves, where a lesion is considered detected if at least one predicted voxel overlaps with the ground-truth region, and case-level confusion matrices that assess whether methods can reliably identify the presence of lesions within each scan. Following established clinical criteria~\cite{topff2025data}, we define a micro-lesion as a connected tumor region with a maximum 3D diameter  $\leq$$3\,\mathrm{mm}$. Higher Dice and AP$_{50}$, and lower HD95 and MAPE, indicate better performance.

\begin{figure}[!t]
\centering
\includegraphics[width=0.9\textwidth]{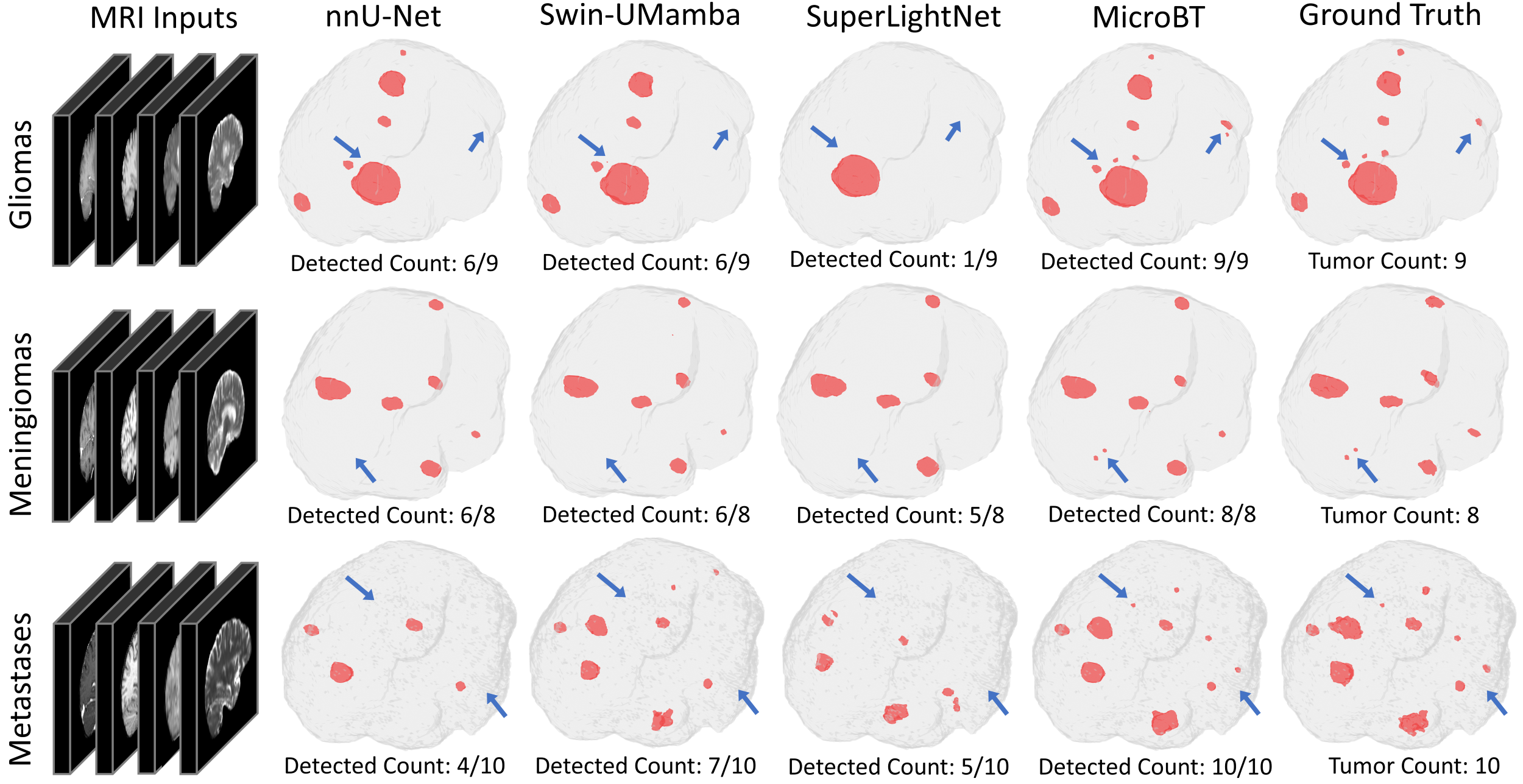}
\caption{Qualitative comparison of segmentation and counting results across different methods. Benefiting from the centroid-anchored synergistic paradigm, the proposed MicroBT framework demonstrates superior micro-lesion sensitivity and more accurate lesion counting compared to state-of-the-art methods.}
\label{fig:res_page2}
\end{figure}

\subsection{Comparison with State-of-the-art Methods}

\noindent \textbf{Micro Brain Tumor Segmentation.} To comprehensively evaluate the segmentation performance of MicroBT, we compare it with nine state-of-the-art methods spanning CNN-based~\cite{isensee2021nnu,milletari2016v}, Transformer-based~\cite{chen2024transunet,zhou2023nnformer,shaker2024unetr++}, and state-space-based~\cite{liu2024swin,ma2024u,xing2025segmamba,gong2025nnmamba} architectures. As shown in Table~\ref{tab:merged_results_v2}, among existing methods, nnU-Net~\cite{isensee2021nnu} achieves the strongest overall baseline with 80.51\% average Dice and 17.01\,mm HD95, while Transformer-based and state-space-based alternatives generally exhibit degraded performance on Metastases, where micro-lesions are most prevalent. For instance, SegMamba~\cite{xing2025segmamba} and nnMamba~\cite{gong2025nnmamba} only achieve 63.45\% and 64.83\% Dice on Metastases, respectively, indicating that efficient long-range modeling alone is insufficient for capturing sub-3\,mm structures. In contrast, our MicroBT consistently outperforms all competitors across three tumor types, achieving 82.12\% average Dice and 13.64\,mm HD95. 

More notably, on Metastases where micro-lesions dominate, MicroBT surpasses the strongest baseline nnU-Net by 2.53\% Dice and 5.67\,mm HD95, and outperforms the best state-space-based method Swin-UMamba~\cite{liu2024swin} by 5.83\% Dice, confirming that the centroid-anchored synergistic paradigm effectively counteracts the systematic suppression of micro-lesions under overlap-based objectives. To further validate statistical reliability, we conduct paired $t$-tests between MicroBT and the strongest baseline nnU-Net across all test cases. As shown in Fig.~\ref{fig:significance}, MicroBT achieves statistically significant improvements ($p < 0.01$) on both Dice and HD95 across all three tumor types, confirming that the observed gains are consistent and not attributable to random variance.

\noindent \textbf{Micro Brain Tumor Counting.} As presented in Table~\ref{tab:detection_counting_results}, we further evaluate the instance-level counting and localization performance. Among existing methods, the dedicated 3D detection method nnDetection~\cite{baumgartner2021nndetection} yields the worst counting reliability with 1.245 average MAPE and only 34.55\% AP$_{50}$, suggesting that generic detection pipelines struggle with the extreme size variation of brain tumors. Segmentation-based methods with post-hoc connected-component counting generally perform better, with nnU-Net~\cite{isensee2021nnu} achieving 0.567 MAPE and 59.06\% AP$_{50}$, and nnMamba~\cite{gong2025nnmamba} reaching 0.549 MAPE and 63.38\% AP$_{50}$. In contrast, our MicroBT achieves the best counting performance across all tumor types, obtaining 0.463 average MAPE and 71.35\% AP$_{50}$. 

On Metastases, where accurate enumeration is most clinically critical for stereotactic radiosurgery planning, MicroBT reduces MAPE from 0.568 to 0.423 and improves AP$_{50}$ from 47.51\% to 65.96\% compared to nnU-Net, representing a 25.5\% relative error reduction and 18.45\% absolute localization gain. We further conduct paired $t$-tests between MicroBT and the strongest counting baseline nnMamba~\cite{gong2025nnmamba}. As shown in Fig.~\ref{fig:significance}, MicroBT achieves highly significant improvements ($p < 0.01$) on both MAPE and AP$_{50}$ for Metastases and Gliomas, and significant improvement ($p < 0.05$) on AP$_{50}$ for Meningiomas. The only non-significant comparison occurs on Meningiomas MAPE, where both methods already achieve low error ($\leq$0.41), leaving limited room for further reduction. Fig.~\ref{fig:res_page2} further provides qualitative comparisons, where MicroBT successfully detects and segments micro-lesions that are missed or merged by competing methods, yielding more accurate lesion counting consistent with ground truth. These results demonstrate that the centroid-guided synergistic learning paradigm enables more reliable instance-level localization than both post-hoc counting and standalone detection approaches. These results demonstrate that the centroid-guided synergistic learning paradigm enables more reliable instance-level localization than both post-hoc counting and standalone detection approaches.

\begin{table*}[!t]
\centering
\caption{Ablation study of our MicroBT framework on Micro Brain Tumor Counting.}
\label{tab:ablation_pyramid}
\resizebox{\textwidth}{!}{%
\begin{tabular}{ccc|cc|cc|cc|cc}
\hline
\multirow{2}{*}{CAS} & \multirow{2}{*}{VC-Loss} & \multirow{2}{*}{SACO} & \multicolumn{2}{c|}{Metastases} & \multicolumn{2}{c|}{Meningiomas} & \multicolumn{2}{c|}{Gliomas} & \multicolumn{2}{c}{Average} \\
\cline{4-5} \cline{6-7} \cline{8-9} \cline{10-11}
 & & & MAPE $\downarrow$ & AP$_{\rm 50}$ $\uparrow$ & MAPE $\downarrow$ & AP$_{\rm 50}$ $\uparrow$ & MAPE $\downarrow$ & AP$_{\rm 50}$ $\uparrow$ & MAPE $\downarrow$ & AP$_{\rm 50}$ $\uparrow$ \\
\hline
~ & ~ & ~ & 5.451 & 53.39 & 0.749 & 53.52 & 0.889 & 70.27 & 2.363 & 59.06 \\
\checkmark & ~ & ~ & 1.385 & 60.00 & 0.454 & 66.20 & 0.678 & 71.81 & 0.839 & 66.00 \\
\checkmark & \checkmark & ~ & 1.082 & 62.63 & 0.420 & 67.06 & 0.574 & 77.93 & 0.692 & 69.21 \\
\checkmark & \checkmark & \checkmark & \textbf{0.423} & \textbf{65.96} & \textbf{0.401} & \textbf{68.52} & \textbf{0.565} & \textbf{79.56} & \textbf{0.463} & \textbf{71.35} \\
\hline
\end{tabular}%
}
\end{table*}

\begin{figure}[!t]
\centering
\includegraphics[width=1\textwidth]{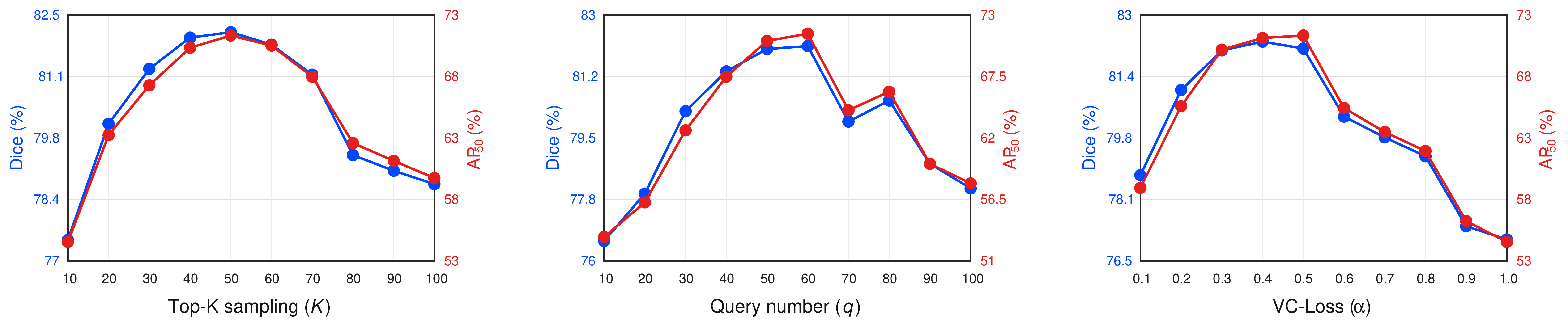}
\caption{Hyperparameter sensitivity analysis on Top-$K$ sampling number, number of object queries $q$, and VC-Loss scaling factor $\alpha$. All three hyperparameters exhibit consistent trends between Dice and AP$_{50}$, with optimal performance achieved at $K$=50, $q$=50, and $\alpha$=0.5.}
\label{fig:hyperparameter}
\end{figure}

\subsection{Ablation Study}

To validate the effectiveness of the proposed Centroid-Anchored Synergistic Paradigm (CAS), Volume-Calibrated Loss (VC-Loss), and Size-Aware Curriculum Optimization (SACO), we conduct comprehensive ablation studies on micro brain tumor counting across three tumor types, as illustrated in Table~\ref{tab:ablation_pyramid}. Starting from the baseline without any proposed component, which yields 2.363 average MAPE and 59.06\% AP$_{50}$, we progressively integrate each module to examine its individual contribution. Introducing CAS brings the most significant improvement, reducing average MAPE from 2.363 to 0.839 and improving AP$_{50}$ from 59.06\% to 66.00\%, confirming that the centroid anchor provides essential spatial guidance for instance-level localization. Incorporating VC-Loss yields an additional 3.21\% AP$_{50}$ gain by amplifying micro-lesion gradient contributions, with particularly notable improvement on Gliomas where AP$_{50}$ increases from 71.81\% to 77.93\%. Integrating SACO further boosts Metastases AP$_{50}$ from 62.63\% to 65.96\%, where micro-lesions are most prevalent, and reduces average MAPE from 0.692 to 0.463, achieving the best average AP$_{50}$ of 71.35\%. These results confirm that CAS, VC-Loss, and SACO collectively contribute to the superior performance of MicroBT, with CAS serving as the foundational enabler and VC-Loss together with SACO providing complementary fine-grained refinement.

Furthermore, we investigate the sensitivity of three key hyperparameters: the Top-$K$ sampling number, the number of object queries $q$, and the VC-Loss scaling factor $\alpha$, as illustrated in Fig.~\ref{fig:hyperparameter}. For Top-$K$ sampling, both Dice and AP$_{50}$ increase steadily from $K$=10, peak around $K$=40$\sim$50, and then gradually decline as excessive candidates introduce false-positive centroids that interfere with downstream branches. A similar trend is observed for $q$, where performance peaks at $q$=50$\sim$60 before degrading, indicating that an insufficient number of queries fails to cover all lesion instances while an excessive number introduces redundant queries that dilute attention on true lesions. For the VC-Loss scaling factor $\alpha$, performance reaches its optimum at $\alpha$=0.4$\sim$0.5, where micro-lesion gradient amplification is sufficient to counteract size imbalance without destabilizing the overall optimization. Beyond $\alpha$=0.6, both metrics drop sharply as overly aggressive weighting causes the training to overfit micro-lesions at the expense of larger tumors. Notably, all three hyperparameters exhibit a consistent bell-shaped pattern between Dice and AP$_{50}$, suggesting that segmentation and counting benefit from the same operating point, which further validates the synergistic design of our centroid-anchored paradigm.

\begin{table*}[!t]
\centering
\caption{Size-stratified comparison on Micro Brain Tumor Counting (MAPE $\downarrow$).}
\label{tab:micro_counting_results}
\tabcolsep=0.13cm
\resizebox{\textwidth}{!}{%
\begin{tabular}{l|cc|cc|cc|cc}
\hline
\multirow{2}{*}{Method} & \multicolumn{2}{c|}{Metastases} & \multicolumn{2}{c|}{Meningiomas} & \multicolumn{2}{c|}{Gliomas} & \multicolumn{2}{c}{Average} \\
\cline{2-3} \cline{4-5} \cline{6-7} \cline{8-9}
 & $\leq$3\,mm & $\leq$5\,mm & $\leq$3\,mm & $\leq$5\,mm & $\leq$3\,mm & $\leq$5\,mm & $\leq$3\,mm & $\leq$5\,mm \\
\hline
nnDetection~\cite{baumgartner2021nndetection} & 1.568 & 1.176 & 2.466 & 2.376 & 1.565 & 1.315 & 1.866 & 1.622 \\
nnU-Net~\cite{isensee2021nnu} & 0.896 & 0.663 & 0.933 & 0.852 & 0.931 & 0.745 & 0.920 & 0.753 \\
TransUNet~\cite{chen2024transunet} & 1.244 & 0.958 & 3.067 & 2.975 & 1.009 & 0.908 & 1.773 & 1.614 \\
nnFormer~\cite{zhou2023nnformer} & 0.950 & 0.703 & 1.667 & 1.589 & 1.000 & 0.780 & 1.206 & 1.024 \\
Swin-UMamba~\cite{liu2024swin} & 0.687 & 0.467 & 3.267 & 3.178 & \textbf{0.824} & \textbf{0.659} & 1.593 & 1.435 \\
UMamba~\cite{ma2024u} & 0.896 & 0.735 & 3.067 & 2.987 & 0.931 & 0.791 & 1.631 & 1.504 \\
UNETR++~\cite{shaker2024unetr++} & 0.864 & 0.596 & 2.000 & 1.906 & 0.941 & 0.696 & 1.268 & 1.066 \\
SegMamba~\cite{xing2025segmamba} & 0.931 & 0.763 & 3.267 & 3.185 & 0.874 & 0.787 & 1.691 & 1.578 \\
nnMamba~\cite{gong2025nnmamba} & 0.955 & 0.707 & 0.698 & 0.626 & 1.039 & 0.779 & 0.897 & 0.704 \\
SuperLightNet~\cite{yu2025superlightnet} & 0.891 & 0.606 & 3.133 & 3.044 & 0.860 & 0.636 & 1.628 & 1.429 \\
\hline
MicroBT (Ours) & \textbf{0.665} & \textbf{0.452} & \textbf{0.667} & \textbf{0.590} & 0.959 & 0.729 & \textbf{0.764} & \textbf{0.591} \\
\hline
\end{tabular}%
}
\end{table*}

\subsection{Size-Stratified Micro-Lesion Analysis}

To further investigate the sensitivity of different methods to micro-lesions of varying scales, we conduct a size-stratified analysis by separately evaluating counting performance at two clinically motivated thresholds: $\leq$3\,mm, which represents the most challenging micro-lesions that are frequently missed in clinical practice, and $\leq$5,mm, which covers small lesions at risk of under-treatment in stereotactic radiosurgery planning~\cite{topff2025data}. As shown in Table~\ref{tab:micro_counting_results}, all methods exhibit notable performance degradation when moving from the $\leq$5\,mm to the $\leq$3\,mm regime, confirming that sub-3\,mm lesions remain the most challenging subset. Among existing methods, several exhibit severe instability on specific tumor types: TransUNet~\cite{chen2024transunet}, Swin-UMamba~\cite{liu2024swin}, and SegMamba~\cite{xing2025segmamba} all exceed 3.0 MAPE on Meningiomas at the $\leq$3\,mm scale, indicating near-complete failure in enumerating the smallest lesions. In contrast, nnU-Net~\cite{isensee2021nnu} and nnMamba~\cite{gong2025nnmamba} demonstrate more balanced performance across tumor types, achieving 0.920 and 0.897 average MAPE at the $\leq$3\,mm scale, respectively. Our MicroBT achieves the best average performance at both size thresholds, obtaining 0.764 and 0.591 MAPE for $\leq$3\,mm and $\leq$5\,mm, respectively, representing a 17.0\% and 16.1\% relative reduction over the strongest baseline nnMamba. On Metastases, where micro-lesions are most clinically significant, MicroBT attains 0.665 MAPE at the $\leq$3\,mm scale, outperforming nnU-Net~\cite{isensee2021nnu} by 25.8\%. These results confirm that the volume-calibrated loss and size-aware curriculum effectively preserve micro-lesion sensitivity across different size regimes, rather than merely improving overall counting statistics.

\begin{figure}[!t]
\centering
\includegraphics[width=0.9\textwidth]{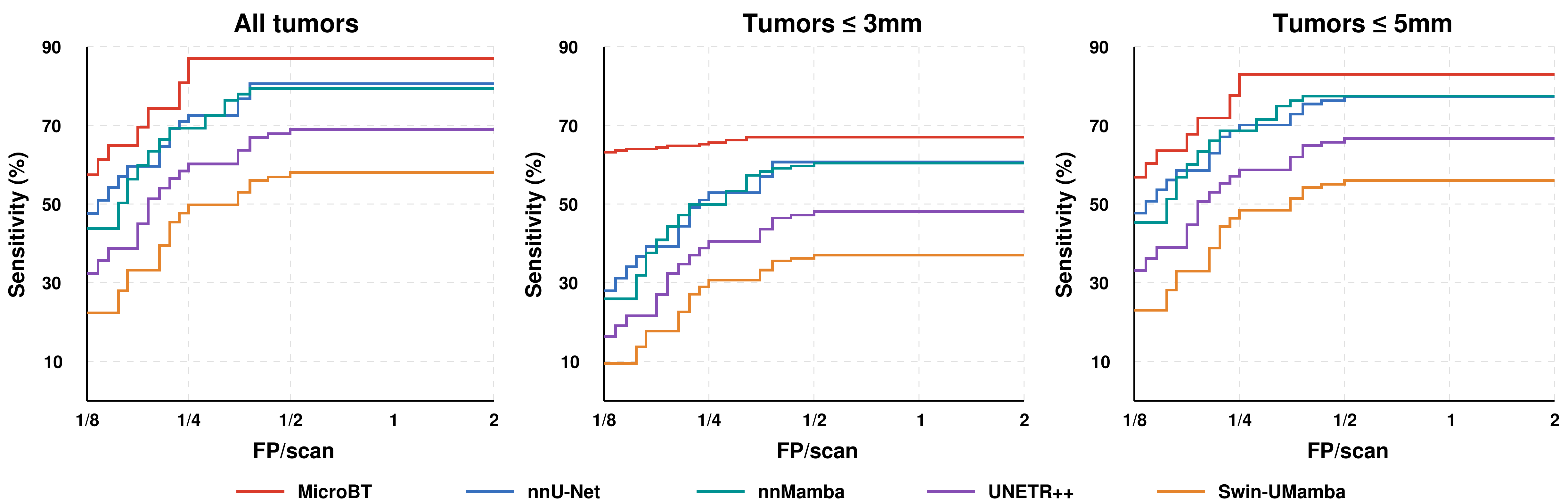}
\caption{Object-level FROC curves across three size regimes. MicroBT achieves the highest sensitivity at all false-positive rates. On the $\leq$3\,mm subset, MicroBT maintains $\sim$68\% sensitivity while competitors plateau below 55\%$\sim$70\%.}
\label{fig:froc}
\end{figure}

To further evaluate clinically relevant detection performance, we report object-level Free-Response Receiver Operating Characteristic (FROC) curves following the evaluation protocol of~\cite{topff2025data}, where a lesion is considered detected if at least one predicted voxel overlaps with the ground-truth region. As shown in Fig.~\ref{fig:froc}, MicroBT consistently achieves the highest sensitivity at all false-positive rates across three size regimes. On the most challenging $\leq$3\,mm subset, existing methods exhibit substantial sensitivity degradation: nnU-Net~\cite{isensee2021nnu} and nnMamba~\cite{gong2025nnmamba} plateau below 55\% and 70\% sensitivity, respectively, while Swin-UMamba~\cite{liu2024swin} struggles to exceed 40\%. In contrast, MicroBT maintains approximately 68\% sensitivity even at low false-positive rates, demonstrating that the centroid-anchored paradigm provides robust spatial priors for detecting micro-lesions that are otherwise suppressed by overlap-based objectives. On the $\leq$5\,mm and all-tumor regimes, MicroBT further widens the gap, reaching approximately 80\% and 85\% sensitivity, respectively, consistently dominating all competing methods across the entire operating range.

\begin{figure}[!t]
\centering
\includegraphics[width=0.7\textwidth]{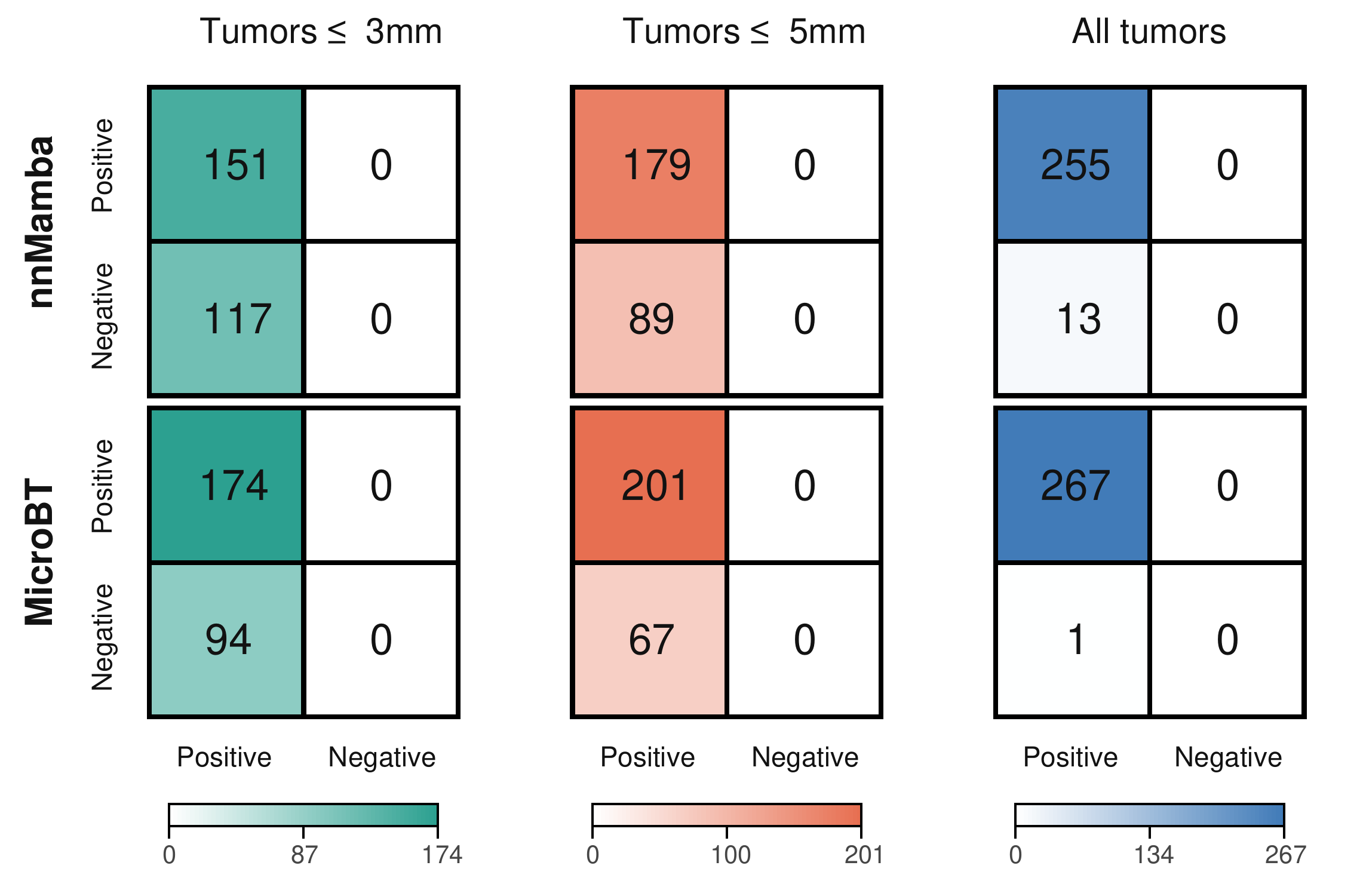}
\caption{Case-level confusion matrices of nnMamba~\cite{gong2025nnmamba} and MicroBT across three size regimes. Note that case counts vary as each threshold yields a different subset of lesion-bearing cases. MicroBT reduces missed cases from 117 to 94 on $\leq$3mm lesions and achieves near-perfect case-level sensitivity on all tumors with only 1 missed case.}
\label{fig:cm}
\end{figure}

We additionally conduct case-level analysis following~\cite{topff2025data} to assess whether methods can reliably identify the presence of lesions within each scan, which directly impacts clinical triage decisions. As shown in Fig.~\ref{fig:cm}, we compare MicroBT against the strongest counting baseline nnMamba~\cite{gong2025nnmamba} via confusion matrices across three size regimes. On the $\leq$3\,mm subset, nnMamba misses 117 out of 268 lesion-bearing cases (43.7\% miss rate), whereas MicroBT reduces the number of missed cases to 94 (35.1\%), recovering 23 additional cases that would otherwise be overlooked. On the $\leq$5\,mm subset, MicroBT further improves case-level sensitivity from 66.8\% to 75.0\%, detecting 22 more lesion-bearing cases than nnMamba. Most notably, on all tumors, MicroBT achieves near-perfect case-level sensitivity with only 1 missed case out of 268 (99.6\%), compared to 13 missed cases by nnMamba (95.1\%). These results highlight that MicroBT not only improves voxel-level segmentation and instance-level counting, but also delivers clinically reliable case-level detection, which is essential for ensuring that no lesion-bearing patient is missed during clinical screening.

\section{Conclusion}
In this work, we identify the critical challenge of micro brain tumor segmentation and counting in multi-modal MRI and propose MicroBT, a centroid-guided synergistic learning framework. Our core insight is that each lesion, regardless of size, possesses a well-defined centroid, which serves as a scale-invariant anchor to bridge dense segmentation and instance-level localization. Built upon this insight, MicroBT devises a CAS that generates centroid heatmaps as shared latent anchors to simultaneously condition a segmentation branch via spatial attention injection and a detection branch via centroid-initialized queries. To counteract the systematic suppression of micro-lesions, we further introduce a VC-Loss that adaptively amplifies gradient contributions of small lesions through volume-normalized exponential weighting, alongside a SACO that stabilizes multi-task training by progressively activating size-aware calibration. Extensive experiments across four datasets demonstrate that MicroBT surpasses state-of-the-art methods across gliomas, meningiomas, and metastases, substantially improving micro-lesion segmentation precision and counting reliability.

\section{Acknowledgments}

This work is partially supported by the Program of China Scholarship Council (202508330191) and PolyU Start-up Fund (P0060371). 

\bibliography{egbib}
\end{document}